\pdfoutput=1

\documentclass[11pt]{article}

\usepackage[preprint]{acl}

\usepackage{times}
\usepackage{latexsym}

\usepackage[T1]{fontenc}

\usepackage[utf8]{inputenc}

\usepackage{microtype}

\usepackage{inconsolata}

\usepackage{graphicx}

\usepackage{multirow}

\usepackage{pgfplots}
\pgfplotsset{compat=1.16}

\usepackage{stfloats}  

\usepackage{booktabs}
\usepackage{titlesec}

\usepackage{siunitx}

\usepackage{makecell}
\usepackage{graphicx}

\title{Low-Resource Cross-Lingual Summarization through Few-Shot Learning with Large Language Models}

\author{Gyutae Park,~Seojin Hwang,~Hwanhee Lee\textsuperscript{$\dagger$}\\
{Department of Artificial Intelligence, Chung-Ang University, Seoul, Korea} \\
\texttt{\{pkt0401, swiftie1230, hwanheelee\}@cau.ac.kr} \\
}

\begin{document}
\maketitle
\footnotetext{\textsuperscript{$\dagger$}Corresponding author.}
\begin{abstract}
Cross-lingual summarization (XLS) aims to generate a summary in a target language different from the source language document. While large language models (LLMs) have shown promising zero-shot XLS performance, their few-shot capabilities on this task remain unexplored, especially for low-resource languages with limited parallel data. In this paper, we investigate the few-shot XLS performance of various models, including Mistral-7B-Instruct-v0.2, GPT-3.5, and GPT-4.
Our experiments demonstrate that few-shot learning significantly improves the XLS performance of LLMs, particularly GPT-3.5 and GPT-4, in low-resource settings. 
However, the open-source model Mistral-7B-Instruct-v0.2 struggles to adapt effectively to the XLS task with limited examples. 
Our findings highlight the potential of few-shot learning for improving XLS performance and the need for further research in designing LLM architectures and pre-training objectives tailored for this task. We provide a future work direction to explore more effective few-shot learning strategies and to investigate the transfer learning capabilities of LLMs for cross-lingual summarization.
\end{abstract}
\section{Introduction}
Cross-Lingual Summarization (XLS) is a task that involves generating a summary in a target language different from the source document's language. It is a complex natural language processing task that requires performing both summarization and machine translation simultaneously. This task is much more challenging than single-language summarization, as it involves overcoming the differences between languages while effectively extracting and compressing key information.
Generally, there are two types of pipelines for XLS systems ~\cite{10.1145/979872.979877,2436/622464}:  \textit{summarize-then-translate} and \textit{translate-then-summarize}. The former summarizes the source text first and then translates it, while the latter translates the source text first and then summarizes it.

\begin{figure}[t]
  \centering
  \includegraphics[width=\columnwidth]{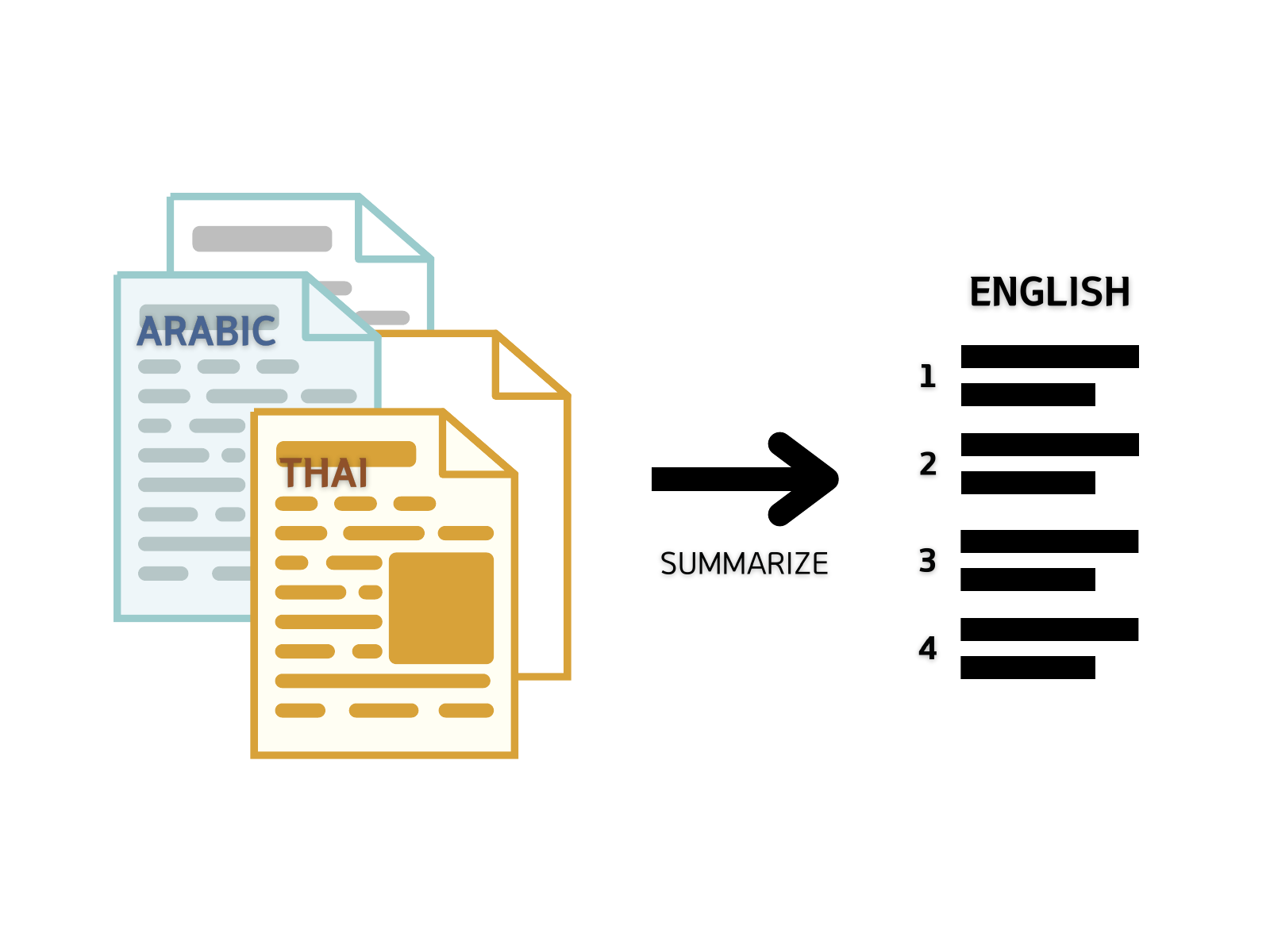}
  \caption{An example of cross-lingual summarization.}
\vspace{-5mm}
\end{figure}

However, these pipelines have the disadvantage that errors occurring at each stage can propagate and accumulate, potentially degrading the final performance.
To resolve the issues of these pipelines, there has been a lot of research on the end-to-end approach~\cite{zhu2019ncls,bai2021crosslingual}, known as the direct method, which generates the target language summary directly from the source document in a single step. The direct method can mitigate the error propagation problem compared to traditional pipelines and enable more efficient learning.

In parallel with the development of end-to-end approaches, Large Language Models (LLMs) such as GPT-3.5 and GPT-4 ~\cite{openai2023gpt4} have demonstrated strong performance in various natural language processing tasks. It is known that these models can significantly improve performance through few-shot learning with only a small number of examples.~\cite{brown2020language}
However, directly applying these models to the XLS task is still challenging. Particularly for low-resource languages, where it is difficult to build large-scale parallel corpora, the lack of data makes it difficult to fully utilize the performance of pre-trained language models, acting as a major obstacle in XLS research.~\cite{ladhak2020wikilingua}
Hence, this study aims to explore the XLS performance of various LLMs using a few-shot learning approach through in-context learning, focusing on the direct method.

Our findings demonstrate that few-shot learning enables LLMs, particularly GPT-3.5 and GPT-4, to achieve competitive performance in cross-lingual summarization tasks for low-resource language settings. The results also highlight open-source models' challenges in adapting to the XLS task with limited parallel data. These findings emphasize the potential of few-shot learning in enhancing cross-lingual summarization capabilities and the need for further research in developing effective few-shot strategies and architectures for low-resource languages.

\section{Related Work}
Cross-lingual summarization (XLS) has undergone significant evolution, shifting from early pipeline approaches like \textit{summarize-then-translate}~\cite{2436/622464,wan-etal-2010-cross} and \textit{translate-then-summarize}~\cite{10.1145/979872.979877,article} to more sophisticated methods utilizing multilingual pre-trained models. These pipelines were initially dominant due to their simplicity but were plagued by error propagation and the limitations inherent to sequential processing tasks. The advent of multilingual pre-trained models such as mBART~\cite{lewis-etal-2020-bart} and mT5~\cite{xue-etal-2021-mt5} marked a transformative shift towards end-to-end approaches, directly generating summaries in the target language and substantially mitigating error propagation issues.

In recent years, the emergence of large language models (LLMs) has revolutionized the field of natural language processing, including XLS. Especially, widely used LLMs like GPT-3.5 and GPT-4 have demonstrated remarkable zero-shot learning capabilities across various tasks.~\cite{brown2020language,qin-etal-2023-chatgpt,bubeck2023sparks} However, the exploration of LLMs in the context of XLS is still in its early stages, with limited research on their zero-shot learning capabilities and even fewer studies focusing on their few-shot learning potential.\cite{wang2023zeroshot}
Recent studies have shown promising results in using LLMs for various NLP tasks.\cite{bang2023multitask,yang2023harnessing,app14052074}

However, the specific exploration of these models in XLS scenarios, particularly in the few-shot setting, remains largely unexplored. While some studies have investigated the zero-shot XLS performance of LLMs~\cite{wang2023zeroshot}, there is a notable lack of research on the few-shot learning capabilities of models such as GPT-3.5, GPT-4, and multilingual LLMs in the XLS domain.
Moreover, the disparity in performance between proprietary models like GPT-4 and open-source alternatives in zero-shot settings underscores the necessity for further investigation into the few-shot capabilities of LLMs. This is particularly critical to ensure that advancements in XLS are equitable and accessible across various linguistic and resource settings.

In this paper, we aim to bridge this gap by exploring the few-shot learning capabilities of LLMs in the context of XLS. We focus on the direct method, leveraging the nuanced capabilities of LLMs like GPT-3.5, GPT-4, and open-source models such as Mistral-7B-Instruct-v0.2. 
The Mistral-7B-Instruct-v0.2 is a 7.3B parameter model that outperforms Llama-2 13B~\cite{touvron2023llama2} across all benchmarks and even surpasses Llama-1 34B~\cite{touvron2023llama1} on many tasks and particularly noted for its ability to process up to 32k tokens, significantly enhancing its capability for few-shot learning by providing richer context management.~\cite{mistralai,jiang2024mixtral} This open-source model has demonstrated strong performance in various natural language processing tasks and offers robust multilingual support. Our aim is to provide comprehensive insights into the practical applications and limitations of these models in low-resource languages, setting the stage for future advancements in the field.

\section{Methods}
The main objective of our research is to compare and analyze the performance of pre-trained mT5 and few-shot prompt-based GPT-3.5 and GPT-4. Then, we aim to experimentally confirm the impact of their few-shot learning on low-resource language XLS tasks. Additionally, we conduct comprehensive comparison with one of the open-source multilingual LLMs, such as the Mistral-7B-Instruct-v0.2 ~\cite{mistralai,jiang2024mixtral}, to provide a broader perspective on the performance of different LLMs in the XLS task to provide a broader perspective on the performance of different LLMs in the XLS task and assess the effectiveness of few-shot learning in mitigating the challenges posed by low-resource settings.

\subsection{Direct Cross-Lingual Summarization}
We focus on the direct cross-lingual summarization method, which generates target language summaries directly from source language documents in an end-to-end manner. Unlike traditional pipelines that involve separate summarization and translation steps, the direct approach combines these tasks into a single, unified process. This allows for a more seamless transfer of information between languages and reduces the potential for error propagation.

\subsection{Models}
We compare the performance of fine-tuned mT5, GPT-3.5, GPT-4, and Mistral-7B-Instruct-v0.2 on cross-lingual summarization tasks. For GPT-3.5 and GPT-4, specifically GPT-3.5-turbo-0125 and GPT-4-0125-preview models, and Mistral-7B-Instruct-v0.2, we evaluate their performance in zero-shot, one-shot, and two-shot settings. The Mistral-7B-Instruct-v0.2 is particularly noted for its ability to process up to 32k tokens, significantly enhancing its capability for few-shot learning by providing richer context management. This model has been shown to outperform other models like Llama-2-13B across all benchmarks, with robust multilingual support enhancing its utility for diverse linguistic datasets.

\begin{figure*}[t] 
  \centering
  \includegraphics[width=\textwidth]{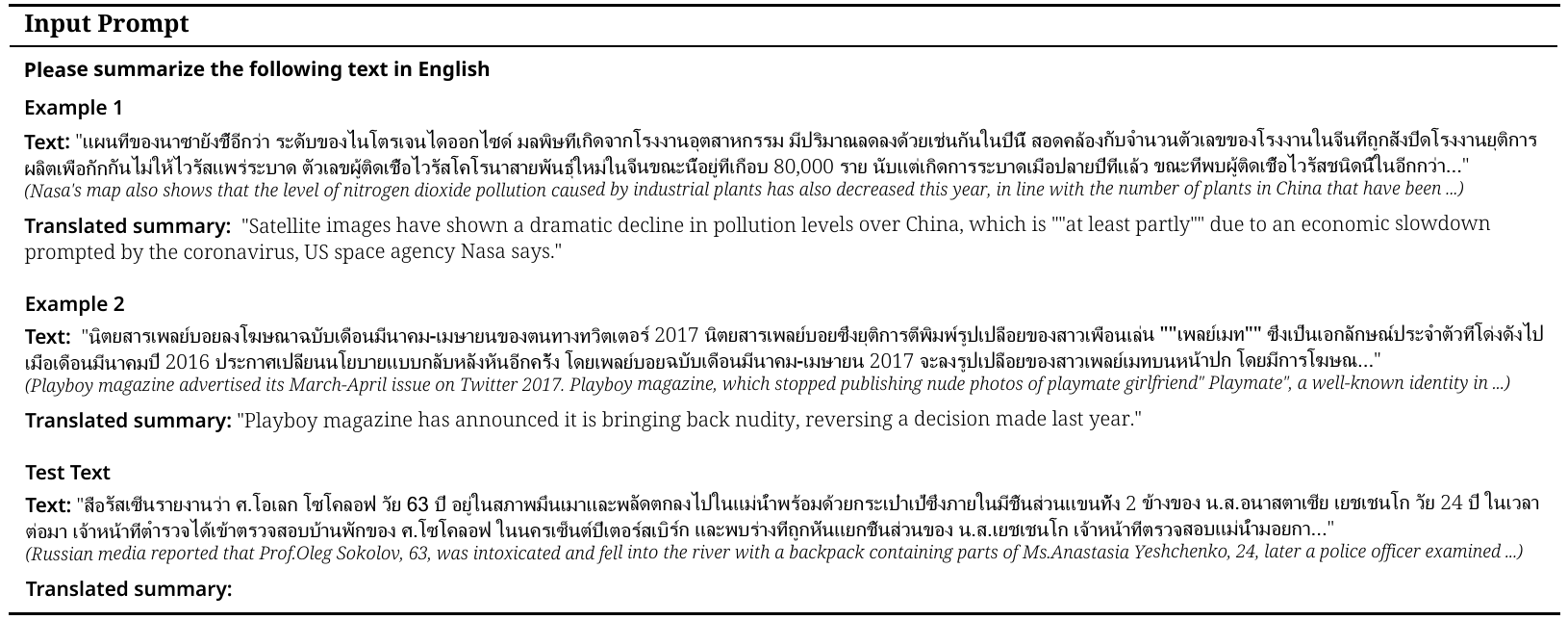} 
  \caption{Two-shot prompt construction for cross-lingual summarization from Thai to English.}
  \label{fig:prompt_example}
\end{figure*}

\subsection{Few-Shot Prompt Construction}
For the few-shot learning approach, we construct prompts that include several examples from the validation set. These examples are carefully selected based on their token count, ensuring that the shortest examples are used as the first and second examples in the prompt. This structured approach facilitates effective few-shot learning, even when computational resources are limited.

The direct prompts for few-shot learning, as depicted in Figure~\ref{fig:prompt_example}, are structured to provide the model with two examples of increasing complexity. Each prompt consists of two example texts and their corresponding summaries in the target language. The test text is then appended to the prompt, and the model is expected to generate a summary in English.
By including these meticulously chosen examples in the prompt, we aim to provide the model with sufficient context to perform few-shot cross-lingual summarization effectively. This method allows us to explore the capabilities of large language models in low-resource settings where the availability of parallel data is limited.

\section{Experiments}

\begin{table*}[t!]
{\LARGE
  \centering
  \setlength{\tabcolsep}{8pt}
  \resizebox{\textwidth}{!}{%
    \begin{tabular}{lSSSSSSSSSSSSSSSSSSS}
        \toprule[3pt]
        \multicolumn{2}{c}{} & \multicolumn{18}{c}{Language Pair} \\
        \cmidrule(rl){3-20}
        \multicolumn{2}{c}{Models} & \multicolumn{3}{c}{Th-En}   & \multicolumn{3}{c}{Gu-En}  & \multicolumn{3}{c}{Mr-En} & \multicolumn{3}{c}{Pa-En}  & \multicolumn{3}{c}{Bu-En} & \multicolumn{3}{c}{Si-En} \\
        
        \cmidrule(rl){3-5} \cmidrule(rl){6-8} \cmidrule(rl){9-11} \cmidrule(rl){12-14} \cmidrule(rl){15-17} \cmidrule(rl){18-20}  
          {}                & {}           & {R-1}           & {R-2}           & {R-L}           & {R-1}           & {R-2}           & {R-L}           & {R-1}           & {R-2}           & {R-L}           & {R-1}           & {R-2}           & {R-L}          & {R-1}           & {R-2}           & {R-L}      
        & {R-1}           & {R-2}           & {R-L}          \\
        \cmidrule[3pt](rl){1-20}
          \multicolumn{2}{c}{mT5-Base (fine-tuning)}      & {29.16}             & {9.79}         & {22.63}          & {24.28}         & {6.40}     & {19.12}          & {25.87}         & {7.28}          & {20.14}         & {30.86}  & {10.27}      & {24.78}             & {28.44}         & {7.50}         & {22.07}             & {28.94}         & {8.17}       & {22.53}              \\
          \cmidrule(rl){1-20}
         {}    &{Zero-shot}       & {14.69}             & {3.84}         & {10.32}          & {12.69}         & {2.80}     & {9.20}          & {13.83}         & {3.04}          & {9.96}         & {14.98}  & {3.78}      & {10.94}             & {13.2}         & {2.21}         & {10.19}             & {13.08}         & {2.11}       & {9.91}              \\
         \cmidrule[0.01pt](rl){2-2}
          \multicolumn{1}{c}{GPT-3.5} &{One\hspace{2em}}       & {15.18}             & {3.95}         & {10.99}          & {13.0}         & {3.12}     & {9.25}          & {14.91}         & {3.73}          & {10.79}         & {14.62}  & {3.93}      & {10.80}             & {15.80}         & {2.43}         & {11.91}             & {13.39}         & {2.34}       & {10.11}              \\
          \cmidrule[0.01pt](rl){2-2}
          {}           &{Two\hspace{2em}}       & {16.56}             & {4.59}         & {11.6}          & {\textbf{15.33}}         & {\textbf{3.57}}     & {\textbf{10.57}}          & {\textbf{16.52}}         & {\textbf{4.06}}          & {\textbf{11.96}}         & {\textbf{15.52}}  & {3.96}      & {\textbf{11.06}}             & {16.48}         & {2.83}         & {\textbf{12.39}}             & {14.03}         & {2.47}       & {9.81}             \\
          \cmidrule(rl){1-20}
          {}    &{Zero-shot}       & {12.22}             & {3.39}         & {8.59}          & {10.96}         & {2.71}     & {7.63}          & {10.97}         & {2.84}          & {7.90}         & {12.10}  & {3.58}      & {8.55}             & {14.73}         & {3.77}         & {9.84}             & {13.92}         & {3.86}       & {9.45}              \\
         \cmidrule[0.01pt](rl){2-2}
          \multicolumn{1}{c}{GPT-4} &{One\hspace{2em}}       & {14.86}             & {4.12}         & {10.51}          & {13.88}         & {3.50}     & {9.27}          & {14.11}         & {3.52}          & {10.09}         & {13.33}  & {4.04}      & {9.43}             & {16.88}         & \textbf{{4.48}}         & {11.41}             & {15.03}         & {\textbf{4.35}}       & {10.27}              \\
          \cmidrule[0.01pt](rl){2-2}
          {}           &{Two\hspace{2em}}       & {\textbf{17.67}}             & {\textbf{5.19}}         & {\textbf{12.67}}          & {13.63}         & {3.49}     & {9.27}          & {14.97}         & {3.73}          & {10.38}         & {13.65}  & {\textbf{4.15}}      & {9.49}             & {\textbf{17.38}}         & {4.32}         & {11.73}             & {\textbf{15.10}}         & {4.15}       & {\textbf{10.39}}              \\
          \cmidrule(rl){1-20}
          \multirow{3}{*}{\makecell{Mistral-7B-\\instruct-v0.2}}    &{Zero-shot}       & {8.91}             & {2.12}         & {6.42}          & {6.15}         & {0.68}     & {4.76}          & {7.63}         & {1.55}          & {5.76}         & {7.65}  & {1.62}      & {5.99}             & {7.59}         & {1.06}         & {5.68}             & {7.23}         & {0.99}       & {5.40}              \\
         \cmidrule[0.01pt](rl){2-2}
         &{One\hspace{2em}}       & {10.28}             & {2.51}         & {7.58}          & {7.27}         & {0.79}     & {5.46}          & {8.08}         & {1.42}          & {6.06}         & {7.41}  & {1.21}      & {5.82}             & {8.35}         & {0.91}         & {6.37}             & {8.43}         & {1.06}       & {6.08}              \\
          \cmidrule[0.01pt](rl){2-2}
          &{Two\hspace{2em}}       & {10.10}             & {2.37}         & {7.30}          & {6.31}         & {0.71}     & {4.81}          & {6.66}         & {0.57}          & {5.35}         & {8.96}  & {1.71}      & {6.77}             & {9.55}         & {1.07}         & {7.44}             & {8.21}         & {0.49}       & {6.58}              \\
        \bottomrule[3pt]
    \end{tabular}
    }\caption{Performance comparison of model performance metrics across various language pairs, including R1, R2,
and Rouge-L scores. The language pairs are abbreviated as follows: Th-En (Thai to English), Gu-En (Gujarati to
English), Mr-En (Marathi to English), Pa-En (Pashto to English), Bu-En (Burmese to English), Si-En (Sinhala to
English).}

  \label{tab:grav4}
  \vspace{-1mm}
}
\end{table*}

\begin{table*}[t!]
{\LARGE
  \centering
  \setlength{\tabcolsep}{8pt}
  \resizebox{\textwidth}{!}{%
    \begin{tabular}{lSSSSSSSSSSSSSSSSSSS}
        \toprule[3pt]
        \multicolumn{2}{c}{} & \multicolumn{18}{c}{Language Pair} \\
        \cmidrule(rl){3-20}
        \multicolumn{2}{c}{Models} & \multicolumn{3}{c}{En-Th} & \multicolumn{3}{c}{En-Gu} & \multicolumn{3}{c}{En-Mr}  & \multicolumn{3}{c}{En-Pa} & \multicolumn{3}{c}{En-Bu} & \multicolumn{3}{c}{En-Si}\\
        \cmidrule(rl){3-5} \cmidrule(rl){6-8} \cmidrule(rl){9-11} \cmidrule(rl){12-14} \cmidrule(rl){15-17} \cmidrule(rl){18-20} 
          {}                & {}           & {R-1}           & {R-2}           & {R-L}           & {R-1}           & {R-2}           & {R-L}           & {R-1}           & {R-2}           & {R-L}           & {R-1}           & {R-2}           & {R-L}          & {R-1}           & {R-2}           & {R-L}      
        & {R-1}           & {R-2}           & {R-L}          \\
        \cmidrule[3pt](rl){1-20}
          \multicolumn{2}{c}{mT5-Base (fine-tuning)}       & {4.59}         & {0.94}             & {4.40}         & {10.14}             & {1.19}         & {9.37}        & {9.38}             & {1.22}         & {8.94}          & {20.12}             & {4.62}         & {17.55}          & {4.85}             & {0.55}         & {4.72}           & {5.62}             & {0.39}         & {5.13} \\
          \cmidrule(rl){1-20}
         {}    &{Zero-shot}       & {4.78}         & {1.48}             & {4.53}         & {1.40}             & {0.0}         & {1.40}        & {2.05}             & {0.0}         & {2.05}          & {0.0}             & {0.0}         & {0.0}          & {1.70}             & {0.26}         & {1.70}           & {1.22}             & {0.0}         & {1.22} \\
         \cmidrule[0.01pt](rl){2-2}
          \multicolumn{1}{c}{GPT-3.5} &{One\hspace{2em}}                   & {4.5}         & {0.82}             & {4.45}         & {1.60}             & {0.0}         & {1.60}        & {1.02}             & {0.13}         & {1.02}          & {0.0}             & {0.0}         & {0.0}          & {0.58}             & {0.0}         & {1.55}          & {1.72}             & {\textbf{1.72}}         & {1.72} \\
          \cmidrule[0.01pt](rl){2-2}
          {}           &{Two\hspace{2em}}                    & {5.29}         & {1.46}             & {5.20}         & {0.55}       & {0.0}         & {0.55}      & {0.88}         & {0.09}        & {0.88}             & {0.0}         & {0.0}          & {0.0}             & {1.55}         & {0.0}          & {1.55}             & {1.63}         & {1.07}           & {1.63}              \\
          \cmidrule(rl){1-20}
          {}    &{Zero-shot}                    & {6.09}         & {1.42}             & {5.84}         & {1.50}             & {0.0}         & {1.50}        & {1.35}             & {0.0}         & {1.35}          & {0.0}             & {0.0}         & {0.0}          & {2.61}             & {0.90}         & {2.61}           & {3.34}             & {\textbf{1.72}}         & {3.34} \\
         \cmidrule[0.01pt](rl){2-2}
          \multicolumn{1}{c}{GPT-4} &{One\hspace{2em}}                    & {\textbf{9.38}}         & {\textbf{2.25}}             & {\textbf{9.38}}         & {1.22}             & {0.0}         & {1.22}        & {1.48}             & {0.1}         & {1.48}          & {0.0}             & {0.0}         & {0.0}          & {4.62}             & {0.17}         & {4.62}           & {\textbf{4.36}}             & {1.15}         & {\textbf{4.36}} \\
          \cmidrule[0.01pt](rl){2-2}
          {}           &{Two\hspace{2em}}                   & {8.64}         & {2.16}             & {8.39}         & {\textbf{2.05}}             & {0.0}         & {\textbf{2.05}}        & {1.31}             & {0.0}         & {1.31}          & {0.0}             & {0.0}         & {0.0}          & {2.02}             & {0.22}         & {2.02}           & {2.53}             & {1.15}         & {2.53} \\
          \cmidrule(rl){1-20}
          \multirow{3}{*}{\makecell{Mistral-7B-\\instruct-v0.2}}    &{Zero-shot}                    &{4.33}         & {1.18}             & {4.33}         & {0.23}             & {0.0}         & {0.23}        & {\textbf{2.37}}             & {\textbf{0.20}}         & {\textbf{2.37}}          & {0.0}             & {0.0}         & {0.0}          & {2.64}             & {1.21}         & {2.64}           & {0.06}             & {0.0}         & {0.06} \\
         \cmidrule[0.01pt](rl){2-2}
         &{One\hspace{2em}}                   & {4.39}         & {1.33}             & {4.30}         & {0.51}             & {0.0}         & {0.51}        & {1.64}             & {0.08}         & {1.64}          & {0.0}             & {0.0}         & {0.0}          & {\textbf{7.78}}             & {\textbf{1.55}}         & {\textbf{7.78}}           & {0.74}             & {0.0}         & {0.74} \\
          \cmidrule[0.01pt](rl){2-2}
          &{Two\hspace{2em}}                    & {3.15}         & {0.38}             & {2.94}         & {0.51}             & {0.0}         & {0.51}        & {0.55}             & {0.0}         & {0.55}          & {0.0}             & {0.0}         & {0.0}          & {2.43}             & {0.01}         & {2.10}           & {0.42}             & {0.0}         & {0.42} \\
        \bottomrule[3pt]
    \end{tabular}
}
    \caption{Performance comparison of model performance metrics across various language pairs, including R1, R2,
and Rouge-L scores. The language pairs are abbreviated as follows: En-Th (English to Thai), En-Gu (English to Gujarati), En-Mr (English to Marathi), En-Pa (English to Pashto), En-Bu (English to Burmese), and En-Si (English to Sinhala).}
  \label{tab:grav4-2}
  \vspace{-5mm}
}
\end{table*}

\subsection{Datasets}
\begin{figure}[h]
\centering
\includegraphics[width=\columnwidth]{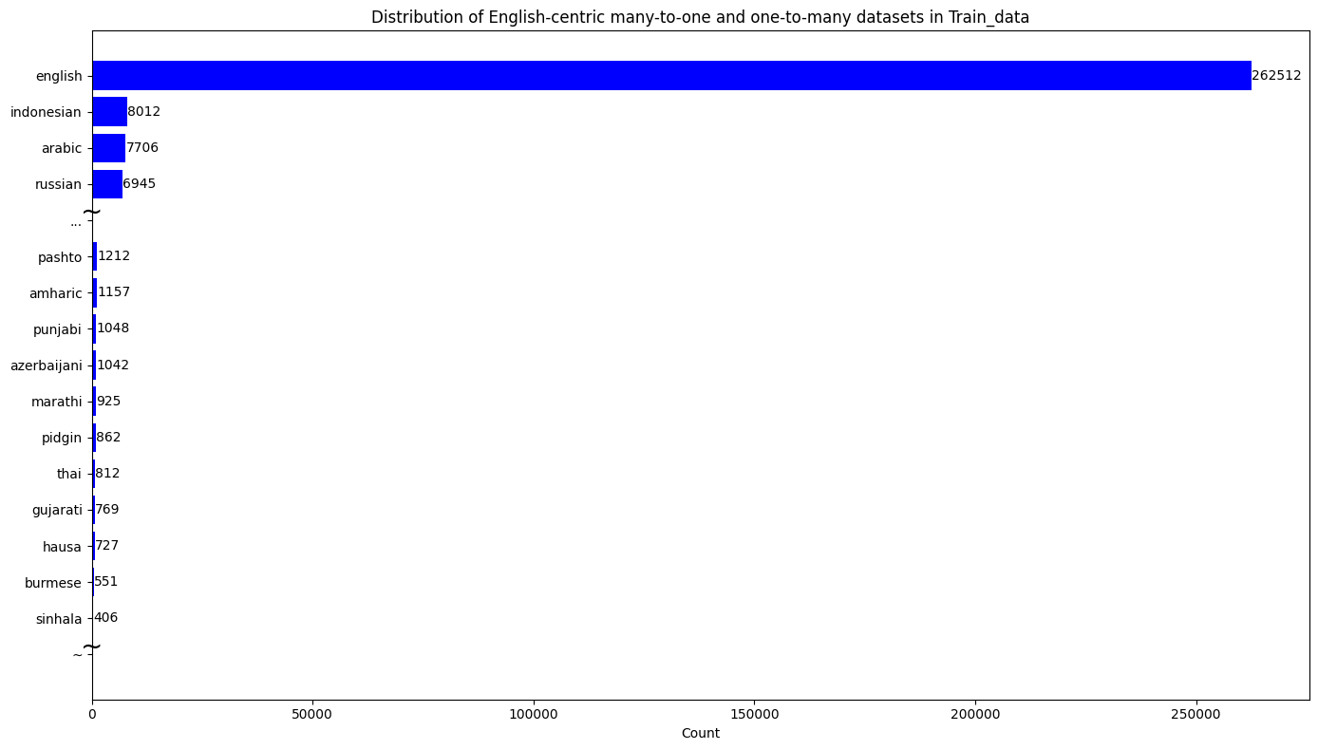}
\caption{Distribution of English-centric many-to-one and one-to-many datasets in Train data}
\label{fig:dataset}
\end{figure}

We utilize the CrossSum~\cite{bhattacharjee2023crosssum} dataset, a multilingual corpus of summaries in 45 languages. Following the definitions in \cite{li-etal-2023-across}, we focus on low-resource language pairs with fewer than 1,000 parallel data points. Additionally, we include experiments with Pashto, a medium-resource language with 1,212 parallel data points, to more broadly assess the effectiveness of our proposed method in diverse linguistic settings.
Figure \ref{fig:dataset} illustrates the distribution of the dataset across different languages for both many-to-one and one-to-many scenarios. The number of parallel data points for each language pair remains consistent in both settings. This symmetry allows us to represent the dataset distribution in a single figure, simplifying the visual representation of the data.

\subsection{Performance Metrics}
To evaluate the quality of the generated summaries, we used the ROUGE ~\cite{lin-2004-rouge}, reporting ROUGE-1/2/L (R-1,2,L). These metrics measure the overlap of unigrams, bigrams, and the longest common subsequences between the generated and reference summaries, respectively.

\subsection{Experimental Results}
\paragraph{Overall Performance:} Fine-tuned mT5 models outperformed most language pairs and experimental settings. Notably, the GPT-3.5 and GPT-4 models demonstrated significant improvements in few-shot scenarios, particularly highlighting their effective adaptation in many-to-one settings, where they summarize from various source languages into English. However, in the one-to-many setting, GPT-3.5, GPT-4, and Mistral-7B-Instruct-v0.2 showed limited performance gains in the one-shot scenario, and their performance either deteriorated or remained unimproved in the two-shot setting. Moreover, there was no significant performance difference among GPT-3.5, GPT-4, and Mistral-7B-Instruct-v0.2 in the one-to-many setting, indicating the challenges associated with summarizing from English to low-resource languages. The Mistral-7B-Instruct-v0.2 model consistently underperformed compared to the fine-tuned mT5 and the GPT-3.5 and GPT-4 models across most language pairs and few-shot settings, suggesting that it struggles to effectively adapt to the cross-lingual summarization task with limited examples.
\paragraph{Few-Shot Learning Impact:}The performance of GPT-3.5 and GPT-4 models competitively improved as the number of shots increased, showcasing their few-shot learning capabilities in cross-lingual summarization. The Mistral-7B-Instruct-v0.2 model exhibited performance gains up to the one-shot setting, with generally increasing ROUGE scores. However, in the two-shot setting, the model's performance showed a decreasing trend, indicating that the benefits of few-shot learning may not consistently extend to higher numbers of shots for this open-source model. This highlights the challenges in applying open-source models to few-shot cross-lingual summarization tasks and suggests that further research is needed to optimize their performance in these settings.
\paragraph{Analysis by Language Pair:}The many-to-one approach generally resulted in higher ROUGE scores than the one-to-many approach. This suggests that summarizing in English is relatively more straightforward than summarizing in other languages. However, the performance gap between the two approaches was more pronounced for the  Mistral-7B-Instruct-v0.2, indicating its limited ability to generate summaries in non-English target languages compared to the other models.
Notably, all models achieved ROUGE scores of 0 for the English to-Pashto language pair across all few-shot settings (see Table~\ref{tab:grav4-2}). This result indicates that few-shot learning did not improve the models' performance for this specific language pair.
The English to Pashto results, where all models failed to generate meaningful summaries even with few-shot learning, underscore the limitations of current approaches in handling extremely low-resource language pairs. This finding emphasizes the need for further research in developing more effective few-shot learning strategies and investigating the transfer learning capabilities of LLMs for cross-lingual summarization in such challenging scenarios.

\section{Conclusion}
This study empirically analyzed the few-shot performance of LLMs in cross-lingual summarization tasks, focusing on low-resource languages using a direct prompting approach. We observed that LLMs demonstrated competitive performance improvements through few-shot learning compared to zero-shot setups particularly in the many-to-one XLS. But, we also demonstrated that there was no significant gain to LLMs in the one-to-many XLS. These findings underscore the need for further research in developing more effective few-shot learning strategies and architectures tailored to low-resource languages.

\section*{Limitation}

Our study conducts experiments on a limited number of low-resource languages and uses only ROUGE metrics to validate the systems' performance. Future research should explore advanced few-shot learning techniques, such as meta-learning or prompt-tuning, and investigate the impact of pre-training objectives and architectures designed specifically for cross-lingual tasks. This could lead to developing more effective open-source models for low-resource cross-lingual summarization.

Despite these limitations, this research demonstrates the potential of large language models' few-shot learning capabilities in low-resource cross-lingual summarization tasks and provides experimental validation for the proposed research directions. Further work is necessary to extend these findings to additional low-resource languages and advance the few-shot learning capabilities of open-source models like Mistral-7B-Instruct-v0.2.

\section*{Acknowledgement}
This research was supported by Institute for Information \& Communications Technology Planning \& Evaluation (IITP) through the Korea government (MSIT) under Grant No. 2021-0-01341 (Artificial Intelligence Graduate School Program (Chung-Ang University)).

\bibliography{custom}

\newpage
\end{document}